\documentclass[conference]{ieeeconf}      

\IEEEoverridecommandlockouts                              

\usepackage{amsmath} 
\usepackage{amssymb} 
\usepackage{amsfonts}
\usepackage{mathtools}
\usepackage{times}  
\usepackage{balance}
\usepackage{graphicx} 
\usepackage{blindtext}
\usepackage[hidelinks]{hyperref}
\usepackage{multirow}
\usepackage{tabularx}
\usepackage{booktabs}
\usepackage{subfig}
\usepackage{float}
\usepackage{color}
\usepackage{cite}
\usepackage[flushleft]{threeparttable}
\usepackage[linesnumbered, ruled]{algorithm2e}

\title{\LARGE \bf
A Robust Pavement Mapping System Based on \\
Normal-Constrained Stereo Visual Odometry}
\author{Huaiyang Huang$^{1*}$, Rui Fan$^{1*}$,~\IEEEmembership{Member,~IEEE}, Yilong Zhu$^{1}$,\\Ming Liu$^{1}$,~\IEEEmembership{Senior Member,~IEEE}, Ioannis Pitas$^2$,~\IEEEmembership{Fellow,~IEEE}\\
\ \ $^{1}$Robotics and Multi-Perception Laboratory, Robotics Institute,\\ the Hong Kong University of Science and Technology, Hong Kong SAR, China. \\
$^{2}$Department of Informatics, Aristotle University of Thessaloniki, Thessaloniki, Greece.\\
Emails: \{hhuangat, eeruifan, yzhubr, eelium\}@ust.hk, pitas@csd.auth.gr\\
\thanks{$^*$These two authors contributed equally to this work and therefore are joint first authors.}
\vspace{-1.5em}
}

\begin{document}

\newcommand\Tstrut{\rule{0pt}{2.6ex}}         
\newcommand\Bstrut{\rule[-0.9ex]{0pt}{0pt}}   

\maketitle
\thispagestyle{empty}
\pagestyle{empty}

\begin{abstract}
    
Pavement condition is crucial for civil
infrastructure maintenance. 
This task usually requires efficient road damage localization, which can be accomplished by the visual odometry system embedded in unmanned aerial vehicles (UAVs).
However, the state-of-the-art visual odometry and mapping methods suffer from large drift under the degeneration of the scene structure. 
To alleviate this issue, we integrate normal constraints into the visual odometry process, which greatly helps to avoid large drift.
By parameterizing the normal vector on the tangential plane, the normal factors are coupled with traditional reprojection factors in the pose optimization procedure. 
The experimental results demonstrate the effectiveness of the proposed system. 
The overall absolute trajectory error is improved by approximately 20\%, which indicates that the estimated trajectory is much more accurate than that obtained using other state-of-the-art methods. 

\end{abstract}

\section{INTRODUCTION}
\label{sec:intro}
Inspecting and repairing pavement damage is an essential task for public infrastructure maintenance \cite{Fan2018}.
Manual visual inspection is still the main form of pavement damage inspection \cite{miller2014distress}, which is, however, very labor-intensive and time-consuming \cite{fan2020pothole}. 
To overcome these drawbacks, more attention is being paid toward developing an automated pavement inspection system \cite{fan2019road}.
However, these solutions are still not robust and precise enough \cite{kim2014review}.
Therefore, how to develop an accurate and efficient pavement damage inspection system is still an open problem.

Recent advances in airborne technology make efficient pavement inspection a more solvable problem \cite{fan2019real}.
Among these techniques, visual odometry (VO) and mapping are two essential modules for an automated pavement inspection system deployed on an unmanned aerial vehicle (UAV).
VO provides UAV systems with a fundamental capability for real-time pose estimation, especially onboard perception sensors \cite{qin2018vins},
while the mapping module allows map establishment and relocalization, and thus global position labeling for pavement damage \cite{mur2017orb}.
Recently, researchers have successfully established autonomous pavement inspection systems for UAVs.
For example, Zhang \textit{et al}. \cite{Zhang2008} designed a robust photogrammetric mapping system for UAVs, which can recognize different types of pavement damages, e.g., cracks and potholes, from RGB images.
Furthermore, Fan \textit{et al}. \cite{fan2019real} proposed an efficient binocular system that is capable of effectively distinguishing road damage from a transformed disparity map  \cite{fan2018novel}.

\begin{figure}[t]
	\centering
	\subfloat[]{\includegraphics[width=0.23\textwidth]{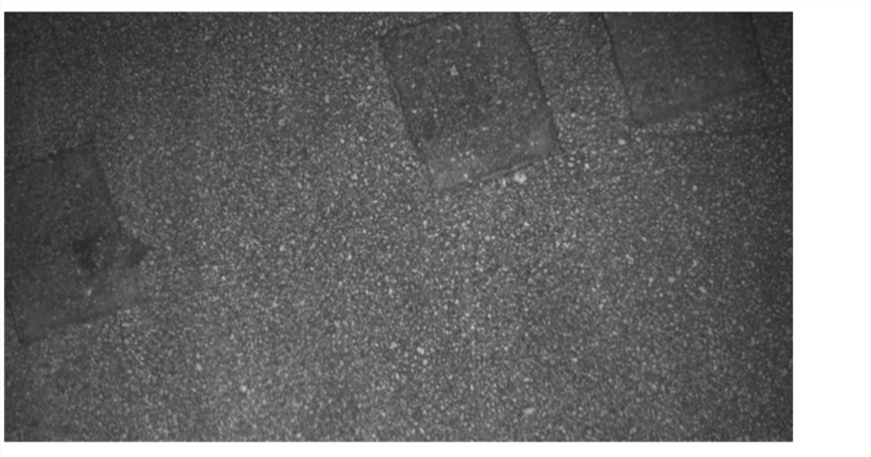}} 
	\subfloat[]{\includegraphics[width=0.23\textwidth]{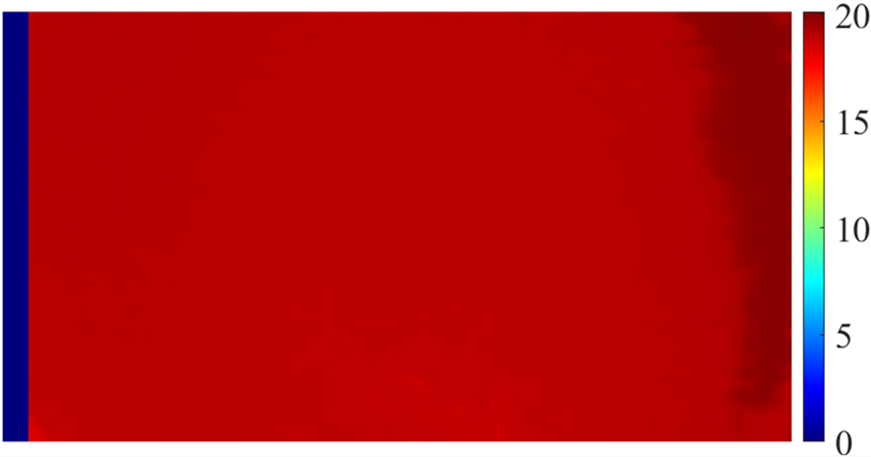}} \\
	\subfloat[\label{fig:system_output}]{\includegraphics[width=0.38\textwidth]{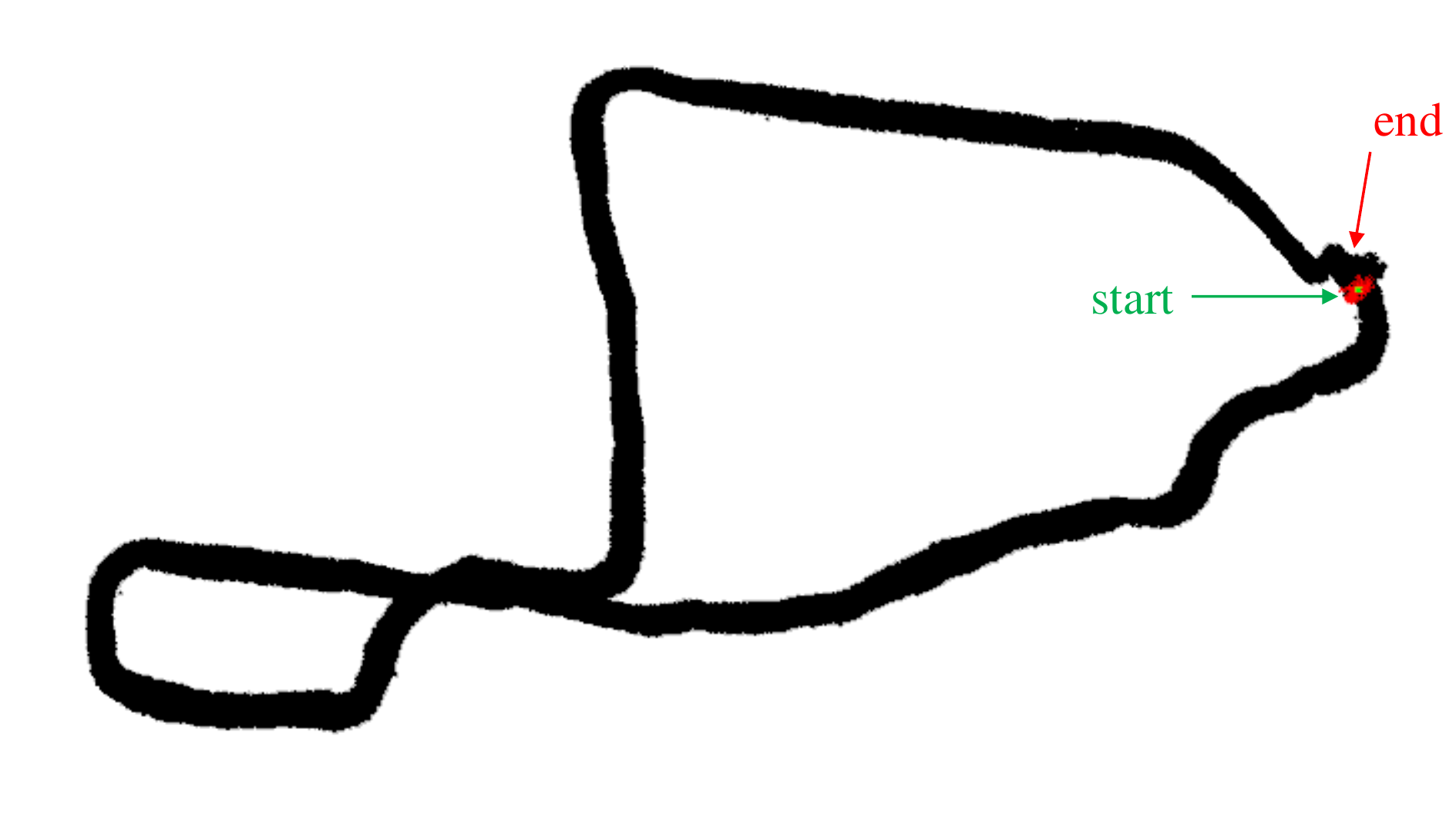}} \\
	\caption{Pavement mapping result under scene degeneration. An input gray-scale image (a) and the corresponding disparity map (b) for a selected keyframe are shown. (c)
		Pavement mapping result produced by the proposed visual odometry and mapping system.}
	\label{fig:camera_input}
	\vspace{-2em}
\end{figure}

Current visual odometry and mapping frameworks have demonstrated their accuracy and robustness on various open-source datasets \cite{engel2018dso, mur2017orb,cvivsic2017soft}.
However, for these state-of-the-art approaches, structure degeneration of visual measurements usually leads to performance degradation in the context of pavement mapping \cite{mur2015orb}.
An example gray-scale image and its corresponding disparity map are shown in \autoref{fig:camera_input}, which shows a near-planar structure and well represents the degeneration issue under this scenario.
To alleviate this problem, we integrate normal constraints into camera pose estimation.
By explicitly minimizing local normal measurements with the global normal prior, we implement a drift-less stereo VO for pavement reconstruction. 
A sample output of the proposed system is shown in Figure 1.(c).

\begin{figure*}
	\centering
	\includegraphics[width=0.95\textwidth]{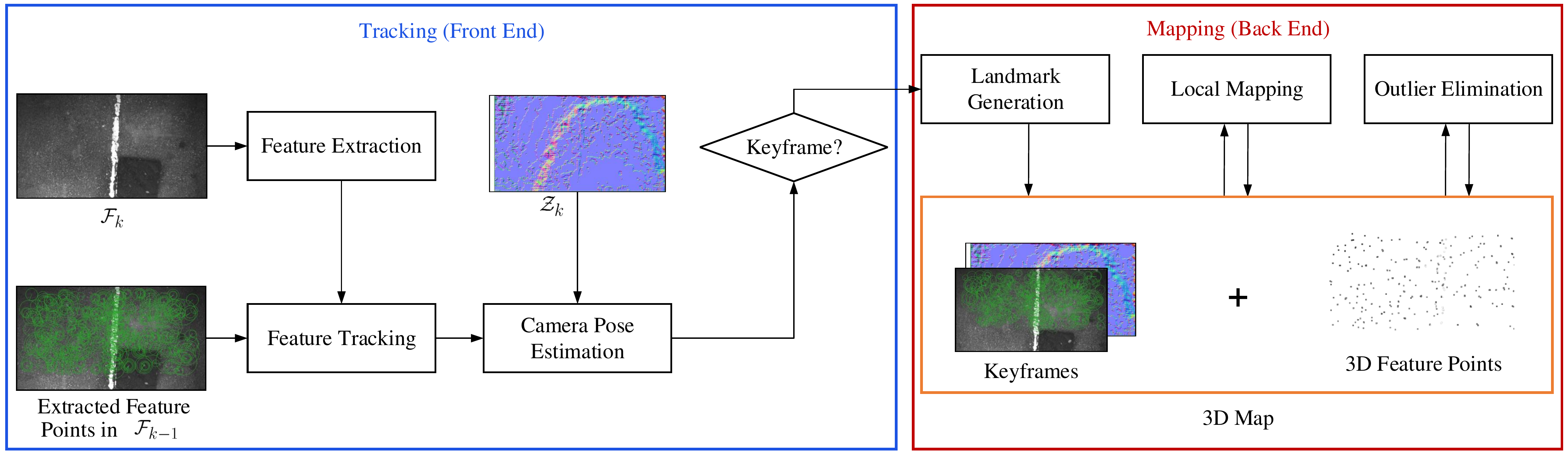}
	\caption{Framework of the proposed visual odometry system.}
	\label{fig:vo_frame}
	\vspace{-1.5em}
\end{figure*}

The remainder of this paper is structured as follows:
\autoref{sec:intro} reviews the state-of-the-art visual odometry methods for UAV pose estimation.
\autoref{sec:method} details the proposed pavement mapping system.
\autoref{sec:experiments} presents the experimental results and discusses the performance of our5 system both qualitatively and quantitatively.
Finally, \autoref{sec:conclusions} summarizes the paper and provides recommendations for future work.
\section{State of the Art}
\label{sec:review}
The state-of-the-art VO or simultaneous localization and mapping (SLAM) systems can be categorized as indirect \cite{mur2015orb, klein2007ptam}, direct \cite{engel2018dso} or hybrid \cite{forster2014svo} methods. 
Recent progress in these methods provides fundamental building blocks for onboard UAV pose estimation both theoretically and technically.
While monocular SLAM algorithms can not recover reliable scale information \cite{fan2019key}.  Recent approaches generally resort to different sensor configurations, including stereo vision or visual-inertial sensors.

Some consider introducing stereo constraints for scale recovery. 
For instance, Cvi{\v{s}}ic \textit{et al}. \cite{cvivsic2017soft} proposed a stereo SLAM framework that is highly efficient in computational demand. To save computational resources, Sun \textit{et al}. \cite{sun2018robust} introduce stereo constraints into a filter-based visual-inertial odometry framework, which was named as S-MSCKF. Raul \textit{et al}. proposed ORB-SLAM2 \cite{mur2017orb}, a versatile visual SLAM framework equipped with sparse mapping, loop-closure and relocalization ability. 

Another common strategy is to fuse visual state estimation with an inertial measurement unit (IMU). This research track tackles the problem based on both filtering and optimization methods. 
Weiss \textit{et al}. \cite{weiss2013monocular} introduced a loosely-coupled filter to recover absolute scale with the aid of an IMU. 6-DoF poses initially estimated by PTAM \cite{klein2007ptam} are fused with the IMU measurements. 
For tightly-coupled filtering, Bloesch \textit{et al}. \cite{bloesch2015robust} extracted multi-level patch features along with 3D landmarks in the camera tracking procedure. Then camera poses are estimated by a standard extended Kalman Filter. 
Leutenegger \textit{et al}.\cite{leutenegger2013keyframe} established a sliding window-based optimization framework with keyframe selection. To optimize camera poses, they formulated a cost function combining both visual reprojection error and inertial error terms.
Forster \textit{et al}. \cite{forster2017manifold} proposed a pre-integration strategy to bootstrap visual-inertial odometry,
while Qin \textit{et al}. \cite{qin2018vins} introduce a loosely-coupled fusion procedure to initialize parameters including scale and bias. 
With IMU measurements pre-integrated, a tightly-coupled back-end jointly optimize camera poses along with other parameters.
\section{Methodology}
\label{sec:method}

\subsection{Notation}
\label{sec:notation}
Throughout the paper, we denote the image collected at the $k$-th time as $I_k$ and the corresponding frame as $\mathcal{F}_k$.
The world coordinate system $\mathcal{F}_w$ is identical to the first camera coordinated system $\mathcal{F}_0$.

For $I_k$, the rigid transform $\mathbf{T}_k \in \mathbf{SE}(3)$ maps a 3D landmark $\mathbf{p} \in \mathbb{R}^3$ to the camera frame using \cite{hartley2003multiple}:
\begin{equation}
\mathbf{p}_k^c = \mathbf{R}_k \mathbf{p}_i + \mathbf{t}_k, 
\end{equation}
where $\mathbf{T}_k = \left[ \mathbf{R}_k | \mathbf{t}_k \right], \mathbf{T}_k \in \mathbf{SE}(3)$. $\mathbf{R}_k$ and $\mathbf{t}_k$ are the rotational and translational components of $\mathbf{T}_k$, respectively.
Accordingly, $\mathbf{p}^c$ denotes a 3D point in $\mathcal{F}_k$.

We use $\pi : \mathbb{R}^3 \rightarrow \mathbb{R}^2$ to denote the projection function: $\mathbf{u} = \pi(\mathbf{p}^c)$, where $\mathbf{u}$ is a pixel in the image coordinate system (ICS). $\pi$ is defined as \cite{fan2018real}:
\begin{equation}
\pi\left(\left[\begin{array}{c}{X} \\ {Y} \\ {Z}\end{array}\right]\right)=\left[\begin{array}{c}{f_{x} \frac{X}{Z}+c_{x}} \\ {f_{y} \frac{Y}{Z}+c_{y}} \\ {f_{x} \frac{X-b}{Z}+c_{x}}\end{array}\right],
\end{equation}
where $(f_x, f_y)$ represents the focal lengths (in pixels) in the horizontal and vertical directions, $(c_x, c_y)$ denotes the principal (in pixels), and $b$ is the baseline of the stereo rig .

The update of a camera pose is parameterized as an incremental twist $\boldsymbol{\xi} \in \mathfrak{se}(3)$. We use a left-multiplicative formulation $\oplus : \mathfrak{se}(3)\times \mathbf{SE}(3)\rightarrow \mathbf{SE}(3)$ to denote the update of $\mathbf{T}_k$, which is denoted as
\begin{equation}
\boldsymbol{\xi}\oplus\mathbf{T}_{k}:=\exp ( \boldsymbol{\xi}^\wedge )\cdot \mathbf{T}_k .
\end{equation}

\subsection{Overview}
\label{sec:overview}



An overview of the proposed system is shown in \autoref{fig:vo_frame}. In the preprocessing stage, for each input frame $I_k$, we compute a dense disparity map and normal map.
Then the relevant information of $I_k$ is delivered to the VO module.
Firstly, the front-end tracker extracts features from accelerated segment test (FAST) \cite{Mair2010} feature points and computes their corresponding Oriented FAST and Rotated BRIEF (ORB) descriptors \cite{Rublee2011}.
With feature association across consequential frames, the current camera pose is estimated through a Perspective-n-Point (PnP) scheme \cite{lepetit2009epnp}.
After initial tracking, the current frame with corresponding matches is delivered to the mapping module, if it is selected as a keyframe.
The back-end optimizer jointly estimates camera poses and 3D positions of landmarks.


\begin{figure}[t!]
	\centering
	\includegraphics[width=0.35\textwidth]{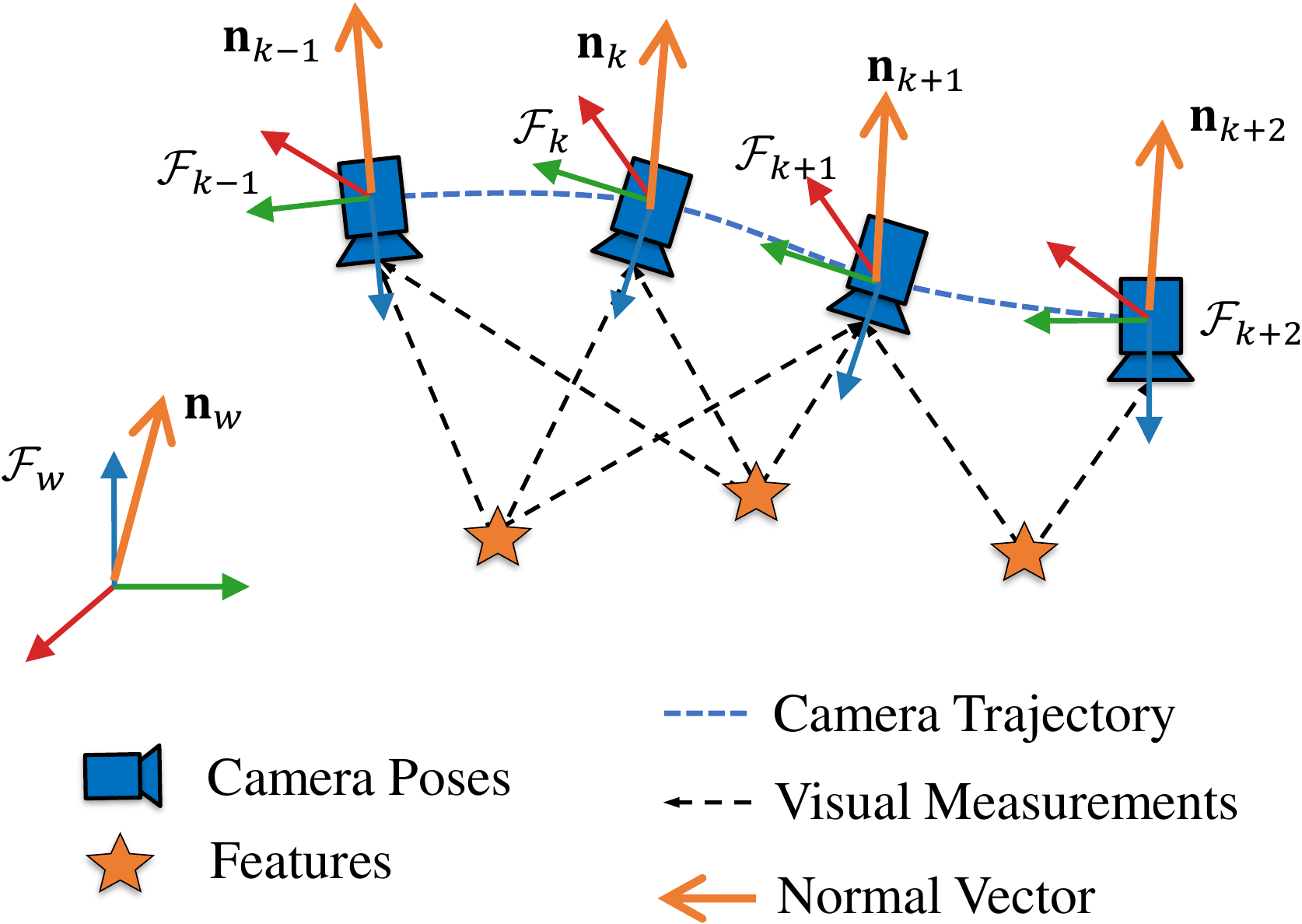}
	\caption{An illustration for camera pose estimation.}
	\label{fig:pose_estimation}
	\vspace{-2em}
\end{figure}

\subsection{Pose Estimation with Normal Constraints}
\label{sec:pose_opt}

This section explains how we introduce the normal constrains into camera pose optimization.
\autoref{fig:pose_estimation} illustrates the Bundle Adjustment (BA) procedure.
We combine two types of factors for camera pose estimation, visual residuals $\mathbf{e}^\text{repro}_{i,k}$ and normal residuals $\mathbf{e}^\text{normal}_k$.

We use the reprojection error as the visual constraint.
The residual term defined on the $i$-th landmark and the $k$-th keyframe is defined as:
\begin{equation}
\mathbf{e}_{i, k}^{\text{repro}} = \pi\left( \mathbf{R}_k \mathbf{p}_i + \mathbf{t}_k \right) - \mathbf{u}_{i, k} ,
\label{eq:residual_normal}
\end{equation}
where $\mathbf{u}_{i, k}$ is the pixel coordinates of the feature associated with $\mathbf{p}_i$ in the $k$-th keyframe.
Generally, the visual constraint measures the distance between the projected position and the observed position.

Although the traditional BA pipeline is able to estimate camera poses, large drift (especially rotation) is observed in the experiments in the context of pavement mapping.
We attribute the failure to strong scene degeneration, as shown in \autoref{fig:camera_input}.
To tackle this issue, we integrate the normal constraints into the pose optimization procedure.
We estimate the normal of the local structure based on the assumption that the 3D geometry is composed of a set of planar surfaces.
Then, by minimizing the residual between global and local surface normals, these factors contribute to the estimation of the rotational component of the camera poses.

\begin{figure}[t!]
	\centering
	\includegraphics[width=0.4\textwidth]{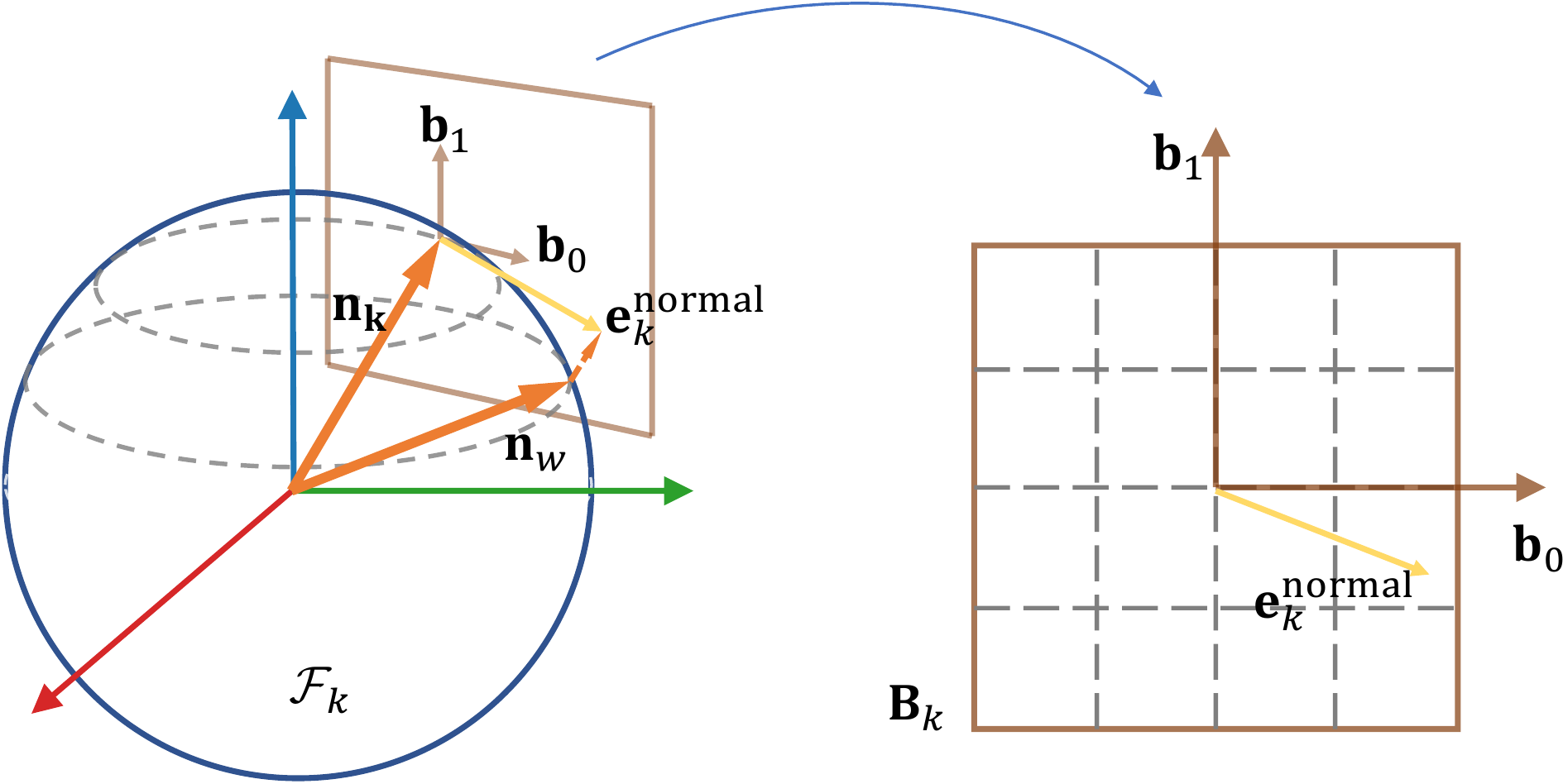}
	\caption{The parameterization of normal vector and corresponding residual.}
	\label{fig:residual}
	\vspace{-1.2em}
\end{figure}

The normal constraints are introduced to provide additional observations under the degenerated scenario.
We leverage the normal-based regulation based on two observations: 1). the pavements in the man-made world typically have a near-planar characteristic;
2). the structure observed in a single frame is a planar surface, which can also be calculated with a closed-form formulation.
We estimate the normal $\mathbf{n}_k$ in each frame.
Inspired by \cite{qin2018vins}, we define this residual term on the tangential plane orthogonal to $\mathbf{n}_k$ under $\mathcal{F}_k$ as:
\begin{equation}
\mathbf{e}_k^{\text{normal}} = \mathbf{B}_k(\mathbf{R}_k \frac{\mathbf{n}_w}{\|\mathbf{n}_w\|} - \mathbf{n}_k),
\label{eq:residual_normal}
\end{equation}
where $\mathbf{B}_k = \left[\mathbf{b}_{k0}, \mathbf{b}_{k1}\right]^T , \mathbf{B}_k \in \mathbb{R}^{2 \times 3}$ consists of two base vectors $\mathbf{b}_{k0}, \mathbf{b}_{k1}$ of the tangential plane.
\autoref{fig:residual} illustrates the definition of the residual terms related to the normal constraints.
From the geometrical perspective, it projects the difference of the global normal and local normal in $\mathcal{F}_k$ onto the tangential plane, yielding a residual term in $\mathbb{R}^2$.
In practice, this avoids overparameterization of normal vectors that can mislead the optimization process.
Additionally, unlike in \cite{qin2018vins}, the optimal normal $\mathbf{n}_k$ observed in $\mathcal{F}_k$ is constant in this formulation.
Therefore, there is no need to recompute $\mathbf{B}_k$ in each iteration, which accelerates the process of optimization.

To generate the two base vectors $\mathbf{b}_{k0}, \mathbf{b}_{k1}$, we arbitrarily assign a unit vector $\mathbf{v}$ that is not parallel to $\mathbf{n}_k$.
Then  $\mathbf{b}_{k0}, \mathbf{b}_{k1}$ can be solved by:
\begin{equation}
\mathbf{b}_{k0}  = \frac{\mathbf{n}_k \times \mathbf{v}}{\|\mathbf{n}_k \times \mathbf{v}\|}, \quad
\mathbf{b}_{k1}  = \frac{\mathbf{n}_k \times \mathbf{b}_{k0}}{\|\mathbf{n}_k \times \mathbf{b}_{k0}\|} .
\end{equation}
The proposed residual definition is more efficient, as there is no need to recompute the bases of the tangential plane when the camera pose is updated.

To estimate the camera pose $\mathbf{T}_k = [\mathbf{R}_k, \mathbf{t}_k]$ of the $k$-th frame, we sum over all the valid factors. Then the camera pose is given by
\begin{equation}
\mathbf{T}_k = \arg \min_{\mathbf{T}_k}  \sum_{i \in \mathcal{P}_{k}} w_{i,k} \|\mathbf{e}_{i, k}^\text{repro}\|_\gamma + \lambda \cdot \|\mathbf{e}^\text{normal}_k\|_\gamma,
\label{eq:pose_opt}
\end{equation}
where $\mathcal{P}_k$ is the set of landmarks matched successfully in the current frame. $\|\cdot\|_\gamma$ stands for the robust Huber norm and $w_{i, k}$ represents the optimization weight associated with pixel variance.
We use a constant factor $\lambda$ to balance the contribution of different factors to the pose optimization.

\subsection{Back-end Optimization}
\label{sec:backend_opt}


The optimization back-end uses a similar objective function as \autoref{eq:pose_opt}.
The parameters to be optimized are denoted as $\mathcal{X}=\{ \mathbf{p}_i, \mathbf{T}_k | i \in \mathcal{P}, k \in \mathcal{T} \}$,
where $\mathcal{P}$ and $\mathcal{T}$ stores the indices of the keyframes and the landmarks in the optimization window, respectively.
We have $\mathcal{P} = \bigcup_{k \in \mathcal{T}} \mathcal{P}_k$
Therefore, a full BA is formulated as:
\begin{equation}
\begin{split}
{\mathcal{X}}
= \arg \min_{\mathcal{X}}
\sum_{k \in \mathcal{T}}
\left[\lambda \cdot \|\mathbf{e}^\text{normal}_k\|_\gamma +
\sum_{i \in \mathcal{P}_k} w_{i,k}
\|\mathbf{e}_{i, k}^\text{repro}\|_\gamma\right].
\end{split}
\end{equation}
Note that, as in \cite{mur2017orb}, the poses of keyframes that do not directly connect to the current keyframe in the covisibility graph are fixed during the optimization.
Like in \cite{mur2017orb}, we detect and reject outliers by $\mathcal{X}^2$-test. Assuming one-pixel variance for every feature, we have $th_{\text{repro}} = 7.815$. For the factor $e_{i,j} > th_\text{repro}$, the corresponding feature $\mathbf{p}_i$ and related observations will be rejected by the mapping module.

To initialize the global normal estimation, we add the global normal into the optimizable parameter set $\mathcal{X}$, yielding an augmented set $\mathcal{X}_\text{init} = \mathcal{X} \bigcup \left\{\mathbf{n}_w \right\}$.
We set a window size $\Delta$ and keep $\mathbf{n}_w$ in the optimizable parameter set until $\Delta$ keyframes have passed to the back-end. Then $\mathbf{n}_w$ is fixed in every subsequent optimization.

\label{sec:exp_setup}
\begin{figure}[t]
	\centering
	\includegraphics[width=0.32\textwidth]{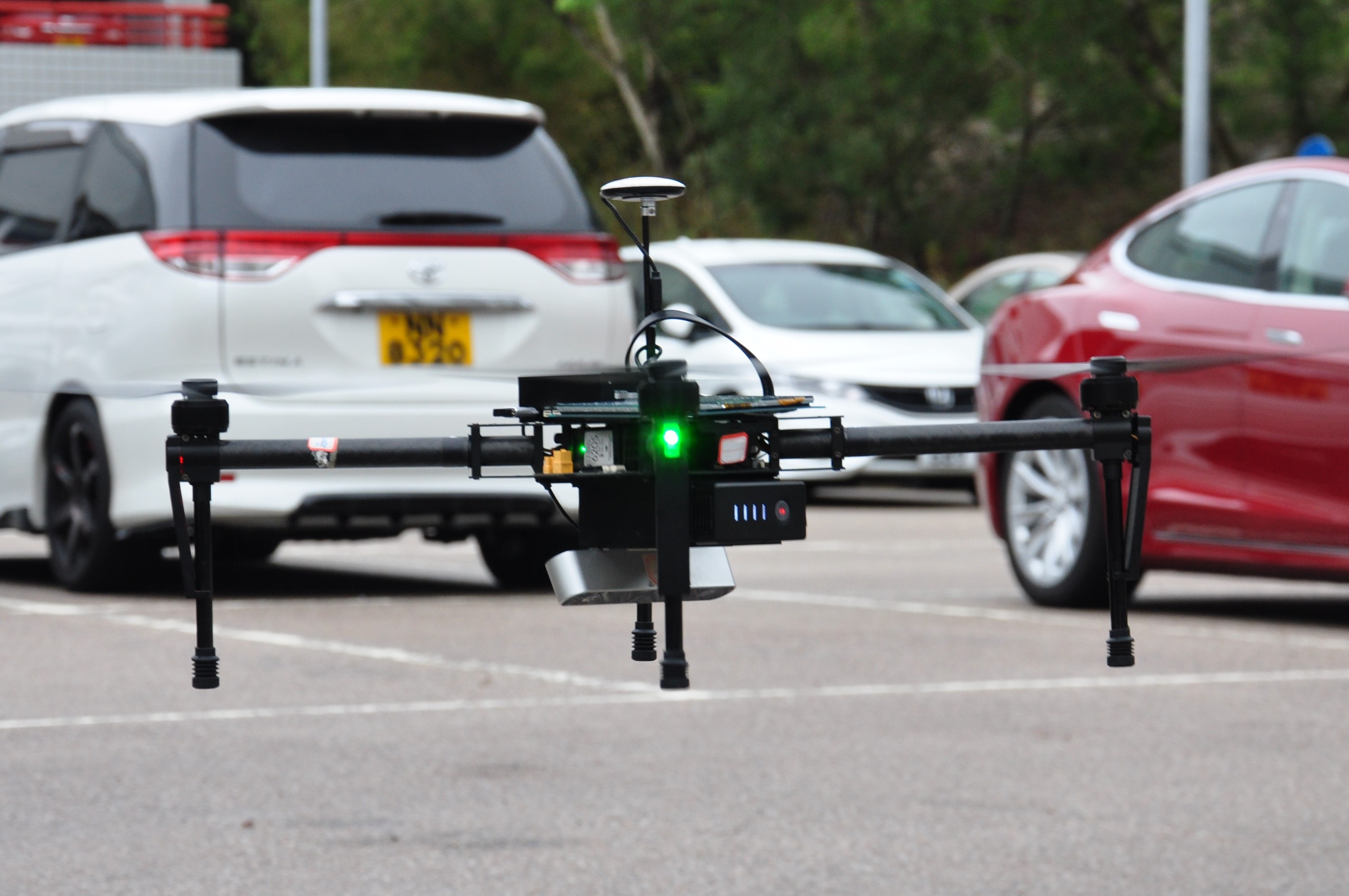}
	\caption{Experimental set-up for acquiring stereo road images.}
	\label{fig:uav_setup}
	\vspace{-1.5em}
\end{figure}

\section{Experimental Results}
\label{sec:experiments}

In this section, we present both qualitative and quantitative experimental results of the proposed normal-constrained stereo VO.
We first describe the experimental set-up and then compare our system with ORB-SLAM2 \cite{mur2017orb} to demonstrate the effectiveness.

\subsection{Experimental Set-Up}

\begin{table*}[htbp]
	\centering
	\caption{ATE of the estimated UAV flight trajectory.}
	\label{tab:ate}
	\begin{tabular}{ccccccccc}
		\hline
		& \multicolumn{4}{c|}{\textbf{ORB-SLAM2}} & \multicolumn{4}{c}{\textbf{Ours}}                                                                                                                                            \\
		Dataset   & \textbf{Mean (m)}                       & \textbf{Median (m)}               & \textbf{RMSE (m)} & \multicolumn{1}{c|}{\textbf{SD (m)}} & \textbf{Mean (m)} & \textbf{Median (m)} & \textbf{RMSE (m)} & \textbf{SD (m)} \\ \hline
		Dataset 1 & 1.657                                   & 1.612                             & 1.790             & \multicolumn{1}{c|}{0.677 }          & 1.462             & 1.268               & 1.657             & 0.779           \\
		Dataset 2 & 4.693                                   & 4.575                             & 5.289            & \multicolumn{1}{c|}{2.436 }          &  2.396                 &     2.311                &         2.632          &        1.090         \\
		Total     & 3.175                                   & 3.094                             & 3.540             & \multicolumn{1}{c|}{1.557 }          &    1.929               &    1.790                 &              2.145     &            0.935     \\  \hline
		\vspace{0.5em}
	\end{tabular}
	\vspace{-1.5em}
\end{table*}

\begin{table*}[htbp]
	\centering
	\caption{RDE  of the estimated UAV flight trajectory.}
	\label{tab:rde}
	\begin{tabular}{ccccccccc}
		\hline
		& \multicolumn{4}{c|}{\textbf{ORB-SLAM2}} & \multicolumn{4}{c}{\textbf{Ours}}                                                                                                                                            \\
		Dataset   & \textbf{Mean (m)}                       & \textbf{Median (m)}               & \textbf{RMSE (m)} & \multicolumn{1}{c|}{\textbf{SD (m)}} & \textbf{Mean (m)} & \textbf{Median (m)} & \textbf{RMSE (m)} & \textbf{SD (m)} \\ \hline
		Dataset 1 & 0.160                                   & 0.109                             & 0.230             & \multicolumn{1}{c|}{0.166 }          & 0.116             & 0.072               & 0.186             & 0.146           \\
		Dataset 2 & 0.159                                   & 0.103                             & 0.244             & \multicolumn{1}{c|}{0.185 }          &         0.119          &     0.071                &    0.205               &       0.167          \\
		Total     & 0.149                                   & 0.106                             & 0.238             & \multicolumn{1}{c|}{0.177 }          &      0.118             &       0.072              &  0.196                 &     0.157            \\  \hline
		\vspace{0.5em}
	\end{tabular}
	\vspace{-1.5em}
\end{table*}

In the experiments, an Intel RealSense  stereo camera D435i\footnote{https://click.intel.com/intel-realsense-depth-camera-d435i-imu.html} was
mounted on a DJI Matrice 100 drone\footnote{https://www.dji.com/uk/matrice100} to acquire stereo road images.
The maximum take-off weight of the drone is 3.6 kg. The stereo camera can capture stereo images with resolution of $1696\times480$ at a speed of 30 fps.
An NVIDIA Jetson TX2 embedded system\footnote{https://developer.nvidia.com/embedded/buy/jetson-tx2} was utilized to save the captured stereo images.  An illustration of the experimental set-up is shown in 
\autoref{fig:uav_setup}. Using the above experimental set-up, four datasets including 26372 stereo image pairs were created. The stereo camera was calibrated manually using the stereo calibration toolbox from MATLAB R2019a. The resolution of the rectified reference and target images is $824\times449$. Also, we mounted a DJI N3 GPS\footnote{https://www.dji.com/hk-en} sensor on the UAV to acquire the flight trajectory ground truth. The GPS precision is $\pm 1 $ m.
The datasets will be publicly available at \url{https://www.ram-lab.com/}.

\subsection{Qualitative Evaluations}
\label{sec:qualtative_res}

\begin{figure}[t!]
	\centering
	\subfloat[]{\includegraphics[width=0.24\textwidth]{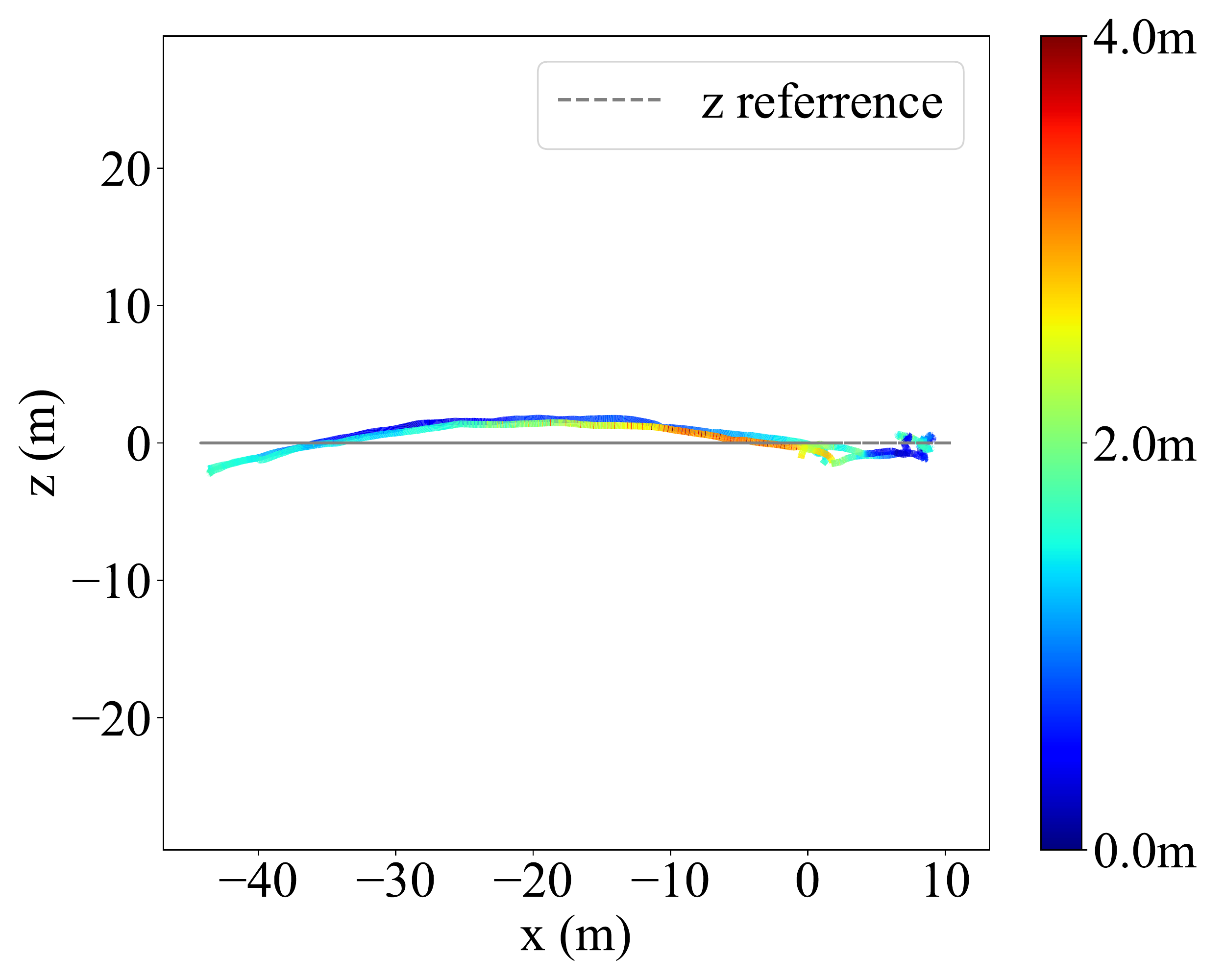}}
	\subfloat[]{\includegraphics[width=0.24\textwidth]{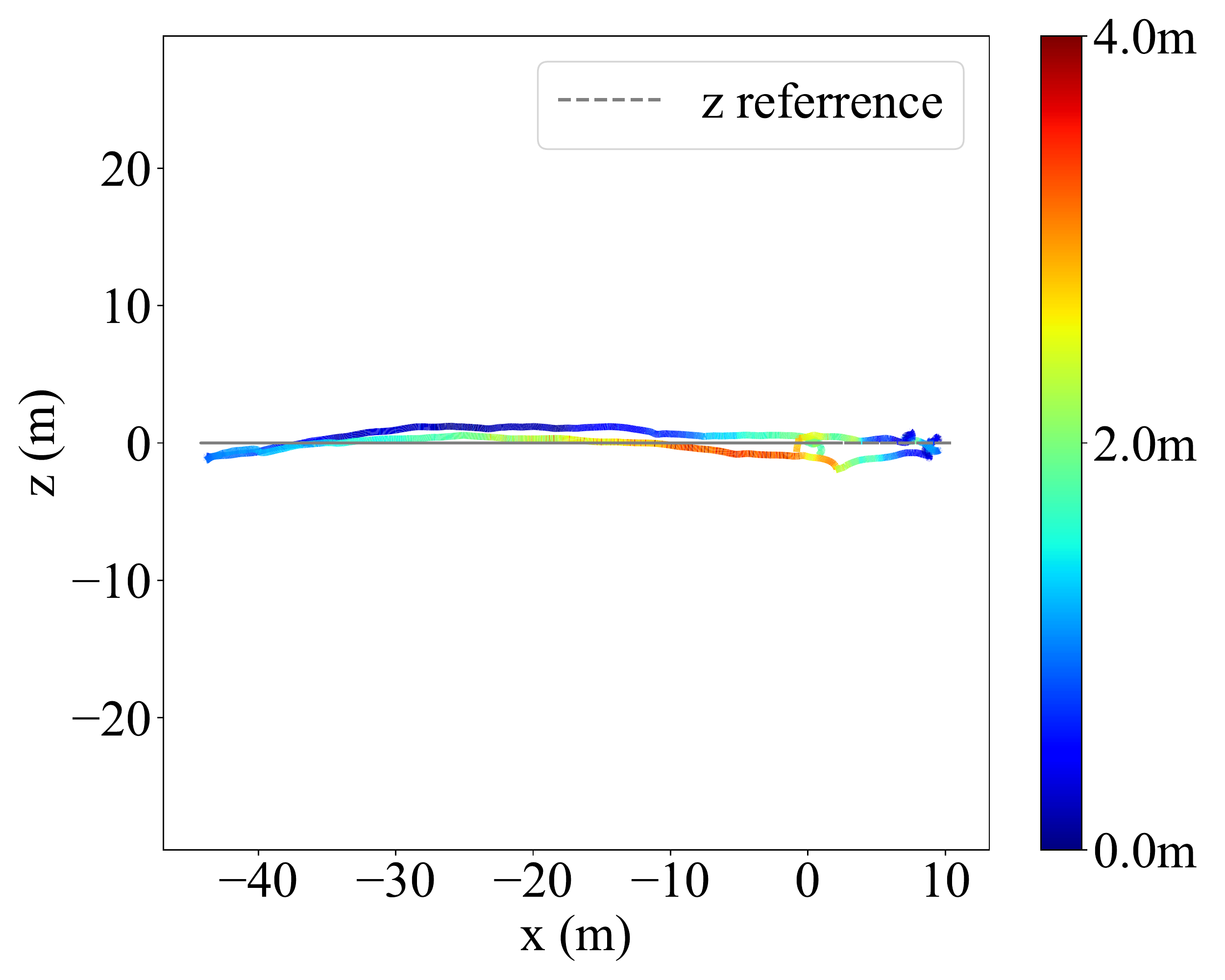}}
	\caption{Comparison between the trajectory estimated by ORB-SLAM2 and ours in the z-axis.}
	\label{fig:wte}
	\vspace{-2em}
\end{figure}

\autoref{fig:wte} compares the proposed system against ORB-SLAM2 \cite{mur2017orb}.
As shown in \autoref{fig:wte}, the odoemtry drift is reduced significantly in the proposed system.
This can be attributed to introducing the normal constraints, so the rotation estimations converge to a better minimum, which benefits the translation estimations in the z-axis.

\subsection{Quantitative Evaluations}
\label{sec:quat_res}

In the experiments, we compare the proposed VO algorithm with ORB-SLAM2 \cite{mur2017orb}, a state-of-the-art stereo visual odometry system.
We compute the absolute trajectory error (ATE) $e^\text{ATE}$ and relative distance error (RDE) $e^\text{RDE}$ between the estimated and ground truth trajectories as the metrics for the evaluations.
$e^\text{ATE}_i$ and $e^\text{RDE}_i$, the ATE and RDE of the $i$-th input frame, are similar to the metrics in \cite{Sturm2012}:
\begin{equation}
e_{i}^\text{ATE} = \Pi(\mathbf{T}_{i}^{-1} \mathbf{S} \mathbf{T}^\text{gt}_{i}),
\end{equation}
\begin{equation}
e_{i}^\text{RDE} =|\Pi\left(\mathbf{T}_{i}^{-1} \mathbf{T}_{i+\Delta}\right)\|_2-\|\Pi\big(({\mathbf{T}^\text{gt}_{i})}^{-1} \mathbf{T}^\text{gt}_{i+\Delta}\big)\|_2|,
\end{equation}
where $\mathbf{T}_{i}$ represents the $i$-th estimated trajectory, $\mathbf{T}^\text{gt}_{i}$ represents the $i$-th ground truth trajectory, $\mathbf{S}$ denotes the rigid-body transformation from $\mathbf{T}^\text{gt}_{i}$ to $\mathbf{T}_{i}$, $\Pi$ is a function to extract the translation components in the $x$- and $y$-axes, and $\Delta=20$ is set to measure the RDE. To quantify the accuracy of the estimated trajectory, we compute the mean, median, root mean square error (RMSE) and standard deviation (SD) of both the ATE and RDE. 

The quantitative results are shown in \autoref{tab:ate} and \autoref{tab:rde}, respectively.
Although ORB-SLAM2 achieves an accurate and consistent camera pose estimation result throughout the experiments, we further improve the accuracy significantly.
According to the comparison, the proposed system generally outperforms ORB SLAM2 and the total ATE is improved .
Note that in Dataset 2, large drift in the z-axis leads to a large ATE and variance of ORB-SLAM2. 
In contrast, the ATE of our method on the same sequence is consistent with the others, which indicates that our method successfully alleviates the VO drift in the z-axis.
Furthermore, the mean and median of the RDE, emphasizing the drift of estimations, of the proposed system are lower than those of ORB-SLAM2. 
This prove the effectiveness of the normal constraints in bounding the drift.

\begin{figure*}[t!]
	\centering
	\subfloat[ORB-SLAM2 on Dataset1]{\includegraphics[width=0.24\textwidth]{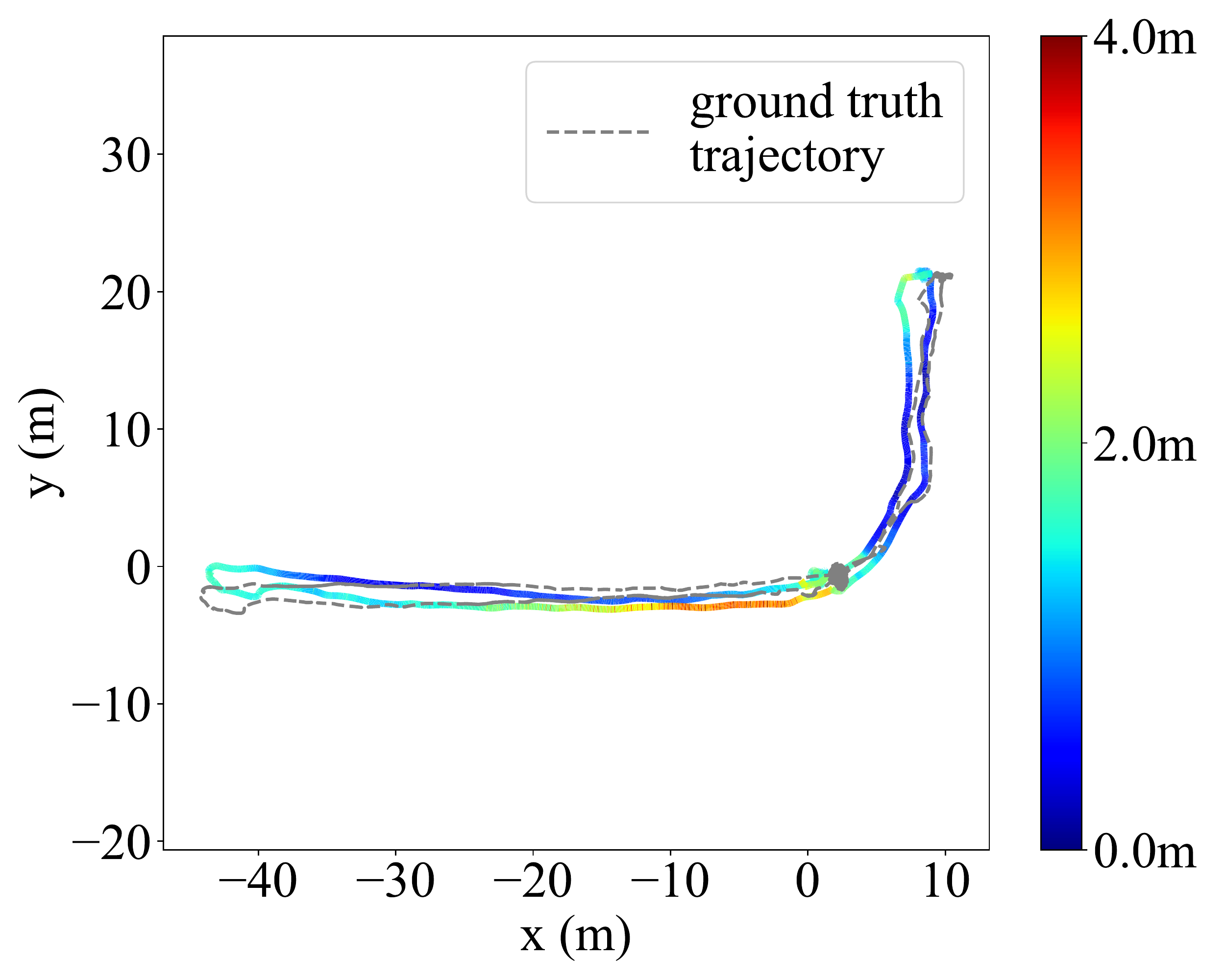}}
	\subfloat[ORB-SLAM2 on Dataset2]{\includegraphics[width=0.24\textwidth]{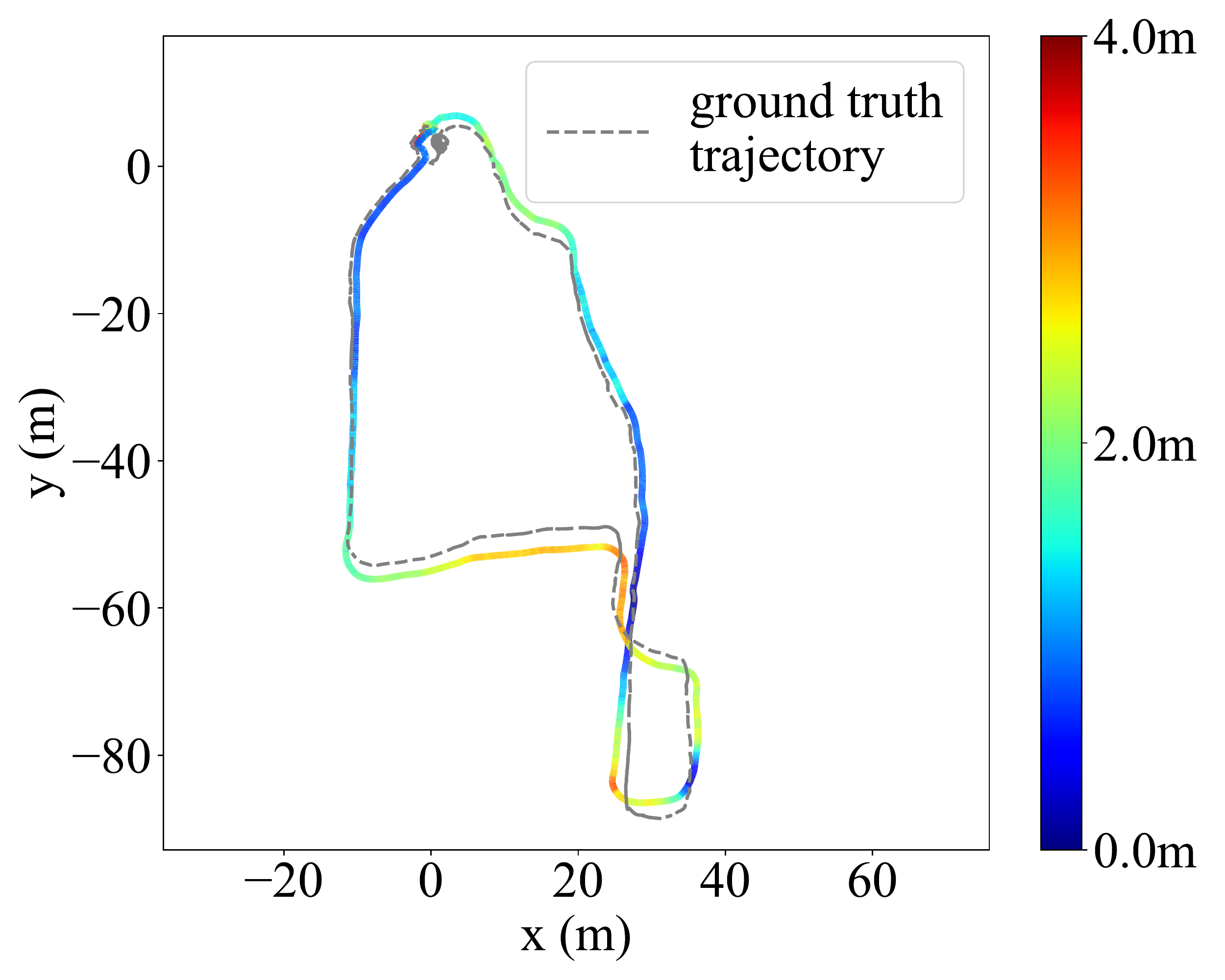}}
	\subfloat[Ours on Dataset1]{\includegraphics[width=0.24\textwidth]{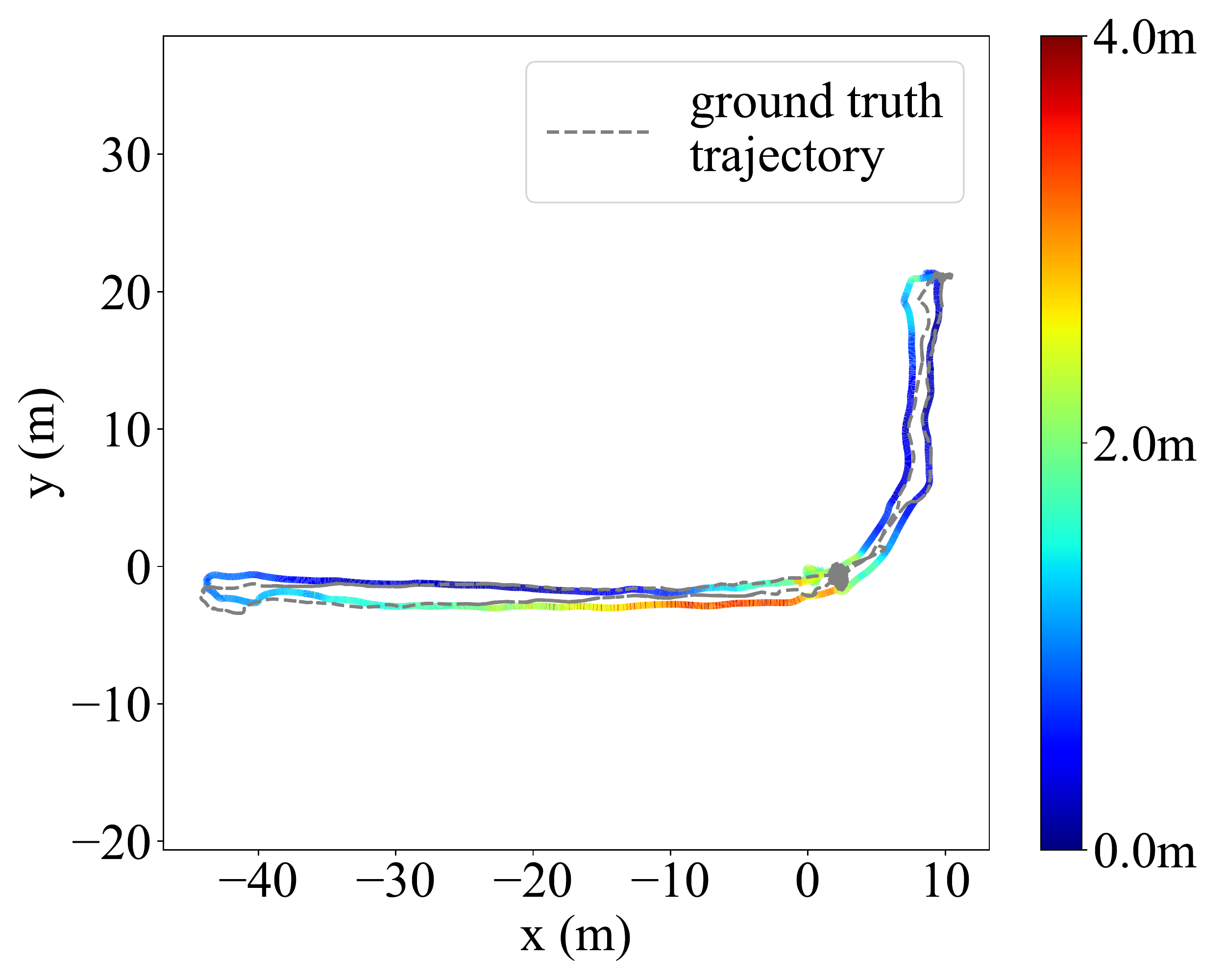}}
	\subfloat[Ours on Dataset2]{\includegraphics[width=0.24\textwidth]{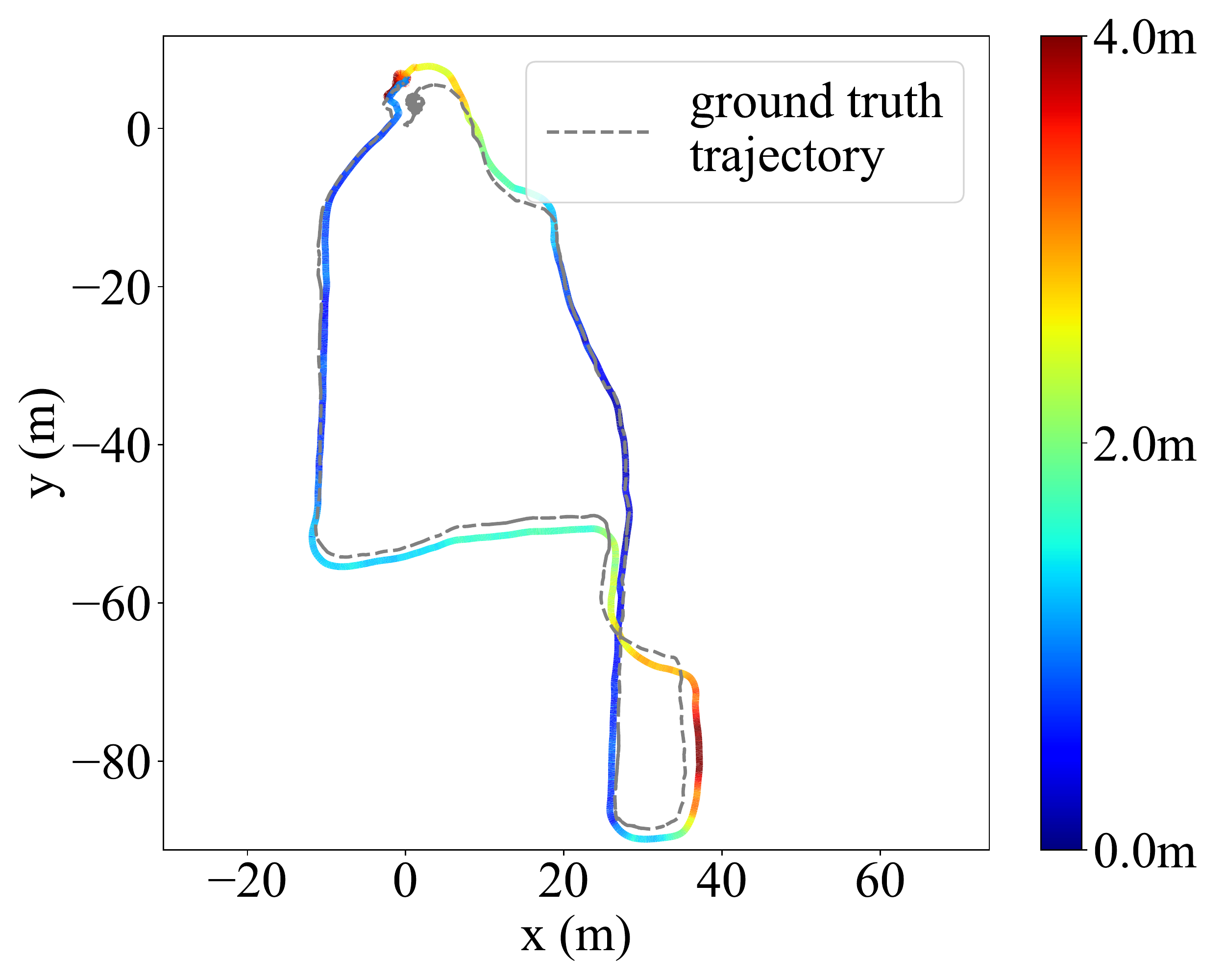}}
	\caption{Comparison between the estimated and ground truth UAV flight trajectory.}
	\vspace{-1em}
	\label{fig:traj_viz}
\end{figure*}

To qualify the performance of the proposed VO system, we align the estimated trajectory with its ground truth, as shown in \autoref{fig:traj_viz}. Our method is better aligned with the ground truth trajectories than ORB-SLAM2.

\subsection{Timing Results}
\label{sec:timing}

\begin{table}[htbp]
	\centering
	\caption{Runtime of the tracking and mapping modules.}
	\label{tab:timing}
	\begin{tabular}{ccccc}
		\hline
		Module   & \textbf{Mean (ms)} & \textbf{Median (ms)} \\ \hline
		Tracking & 28.7               & 26.5                 \\
		Mapping  & 46.8               & 44.5                 \\
		\hline
	\end{tabular}
	\vspace{-1.5em}
\end{table}

We measure the runtime of both the tracking and mapping modules, as shown in \autoref{tab:timing}.
The proposed system achieves a camera tracking speed of approximately 30 fps when running on an Intel Core i7-8700k CPU. 
The timing results demonstrate the real-time performance of the proposed normal-constrained stereo SLAM system.

\section{Conclusions and Future Work}
\label{sec:conclusions}

In this paper, we presented a pavement mapping system that explicitly introduces ground normal estimations as constraints in the pose optimization and bundle adjustment. We discussed an effective parameterization of normal vectors and corresponding residual term in the optimization. The experimental results showed that our method is drift-less and more accurate than the state-of-the-art ones, which demonstrated the effectiveness of coupling normal constraints with traditional bundle adjustment pipeline.

In the future, we will render the structure assumption become more applicable for pavement mapping. 
Through minimizing the error between normal observations of single landmarks, the odometry drift might be reduced without explicitly assuming a global constraints.
Additionally, with binocular vision only, current implementation is not robust enough under rapid motion (as this may cause motion blur). 
Therefore, we are planning to leverage inertial measurements and visual-inertial coupling for a more robust visual odometry system. Furthermore, we plan to use our recently published work \cite{fan2019robust} to provide the UAV with the roll angle information. We believe this can further improve the odometry accuracy. 
\section*{Acknowledgment}
This work was supported by the National Natural Science Foundation of China, under grant No. U1713211, the Research Grant Council of Hong Kong SAR Government, China, under Project No. 11210017, No. 21202816, and the Shenzhen Science, Technology and Innovation Commission (SZSTI) under grant JCYJ20160428154842603, awarded to Prof. Ming Liu.








\bibliography{slam}
\balance


\end{document}